\documentclass{article}

\usepackage{neurips_2021}

\usepackage[utf8]{inputenc} % allow utf-8 input
\usepackage[T1]{fontenc}    % use 8-bit T1 fonts
\usepackage{hyperref}       % hyperlinks
\usepackage{url}            % simple URL typesetting
\usepackage{booktabs}       % professional-quality tables
\usepackage{amsfonts}       % blackboard math symbols
\usepackage{nicefrac}       % compact symbols for 1/2, etc.
\usepackage{microtype}      % microtypography
\usepackage{xcolor}         % colors
\usepackage{xspace}

\newcommand{\method}{\textsc{SEDGE}\xspace}
\newcommand{\methodplus}{\textsc{SEDGE}}

\usepackage{makecell}
\usepackage{amssymb}% http://ctan.org/pkg/amssymb
\usepackage{pifont}% http://ctan.org/pkg/pifont
\usepackage{booktabs}

\usepackage{caption}
\usepackage{graphicx}
\usepackage{float} 
\usepackage{subfigure}

\usepackage{tikz}
\usepackage{comment}
\usepackage{amsmath,amssymb} % define this before the line numbering.
\usepackage{color}
\usepackage{multirow}
\usepackage{bbding}
\usepackage{utfsym}
\usepackage{wrapfig}

\title{Domain Generalization using Pretrained Models without Fine-tuning}

\author{Ziyue Li\\
  ShanghaiTech University\\
  \texttt{lizy@shanghaitech.edu.cn} \\
   \And
   Kan Ren \\
   Microsoft Research Asia \\
   \texttt{kan.ren@microsoft.com} \\
   \AND
   Xinyang Jiang \\
   Microsoft Research Asia \\
   \texttt{xinyangjiang@microsoft.com} \\
   \And
   Bo Li \\
   Nanyang Technological University\\
   \texttt{libo0013@e.ntu.edu.sg} \\
   \And
   Haipeng Zhang \\
   ShanghaiTech University\\
   \texttt{zhanghp@shanghaitech.edu.cn} \\
   \And
   Dongsheng Li \\
   Microsoft Research Asia \\
   \texttt{dongsli@microsoft.com} \\
}

\begin{document}

\maketitle

\begin{abstract}
Fine-tuning pretrained models is a common practice in domain generalization (DG) tasks. 
However, fine-tuning is usually computationally expensive due to the ever-growing size of pretrained models. More importantly, it may cause over-fitting on source domain and compromise their generalization ability as shown in recent works. 
Generally, pretrained models possess some level of generalization ability and can achieve decent performance regarding specific domains and samples. 
However, the generalization performance of pretrained models 
could vary significantly over different test domains even samples, which raises challenges for us to best leverage pretrained models in DG tasks. 
In this paper, we propose a novel domain generalization paradigm to better leverage various pretrained models, named {\bf s}pecialized {\bf e}nsemble learning for {\bf d}omain {\bf ge}neralization (SEDGE). 
It first trains a linear label space adapter upon fixed pretrained models, which transforms the outputs of the pretrained model to the label space of the target domain.
Then, an ensemble network aware of model specialty is proposed to dynamically dispatch proper pretrained models to predict each test sample.
Experimental studies on several benchmarks show that SEDGE achieves significant performance improvements 
comparing to strong baselines including state-of-the-art method in DG tasks 
and reduces the trainable parameters by $\sim99\%$ and the training time by $\sim99.5\%$.

\end{abstract}

\section{Introduction}

Distribution shift is a common problem 
caused by physical or psychological factors of the real-world applications, which breaks the independent and identically distributional (i.i.d.) assumption of machine learning algorithms.
Thus, \textit{generalization} becomes increasingly important when training and applying machine learning models in practice.

The task of domain generalization and the corresponding benchmark ~\citep{gulrajani2020search} have been proposed for studying and improving model generalization by training on source domains and test on target domains.
These methods focus on generalizable model training following a fine-tuning paradigm which often leverages pretrained models like ResNet ~\citep{he2016deep} as an initialization and fine-tunes that with some elaborate training algorithms on the source domains. 
Then, the trained models would be evaluated on the unseen target domains.
One common assumption behind this commonly used paradigm is that fine-tuning brings better performance.
However, fine-tuning is usually computationally expensive due to the ever-growing size of pretrained model, and proven to possibly compromise the generalization ability of pretrained models and under-perform in out-of-distribution scenarios ~\citep{kumar2021fine,yu2021empirical}.

\begin{figure}[t]
    \begin{center}
    \includegraphics[width=\textwidth]{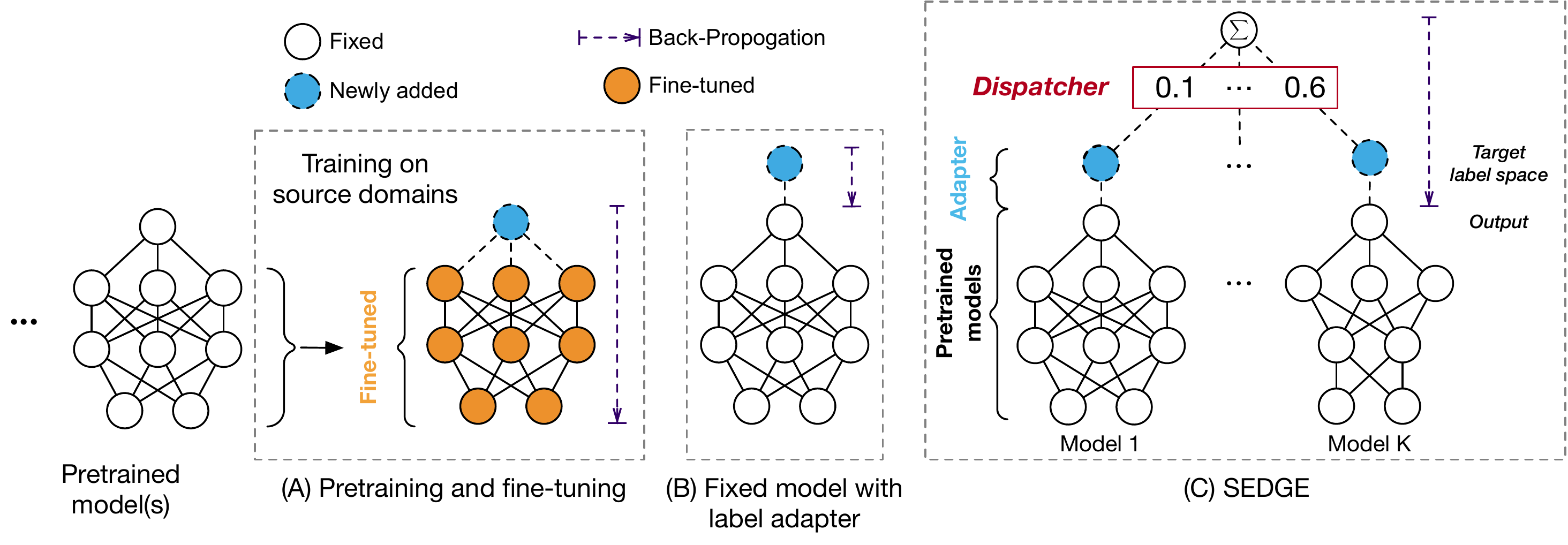}
  \end{center}

  \caption{Different training paradigms in domain generalization.
}
  \label{fig:training_paradigm}

\end{figure}

Therefore, instead of using pretrained model as an initialization like most existing methods on domain generalization, this paper seeks a better way to leverage the vast amount of the existing pretrained models ~\citep{he2016deep,krizhevsky2012imagenet,iandola2014densenet,zoph2018learning,he2021masked,radford2021learning}. 
Generally, the existing pretrained models have already possessed certain generalization ability over out-of-distribution scenarios. 
As shown in Figure \ref{fig:training_paradigm} (B), one simple way to exploit pretrained models' generalization ability is to train a linear label space adapter over a fixed weight pretrained model, which directly transforms the outputs of the pretrained model to the target label space.
Our experiments show this minor adjustment bring enhancement over the fine-tuned one on certain target domains (detailed results in section \ref{sec:ablation}).

However, fixed pretrained models do not constantly generalize on all domains, and the generalization performances of different pretrained models vary significantly over different target domains, label classes or even samples, as shown in Figure~\ref{fig:spec different domains}.
This is caused by various aspects of the pretraining procedure such as model hypothesis, training algorithms and pretraining datasets.

Due to the significant variance of pretrained models' generalization ability, it is essential to identify the samples a pretrained model generalize to (i.e. model specialty). 
Here we propose a novel learning paradigm that dispatches proper pretrained models to each sample based on their generalization ability, named {\bf s}pecialized {\bf e}nsemble learning for {\bf d}omain {\bf ge}neralization (SEDGE). 
As shown in Figure \ref{fig:training_paradigm} (C), specifically, in addition to the label adapter that projects the pretrained domain to the target domain upon the  model with the fixed parameters, we further
incorporate a model specialty aware ensemble network
that selects a set of proper pretrained models 
and aggregate together to conduct predictions for each specific sample.

\begin{figure}
    \begin{center}

    \includegraphics[width=.8\textwidth]{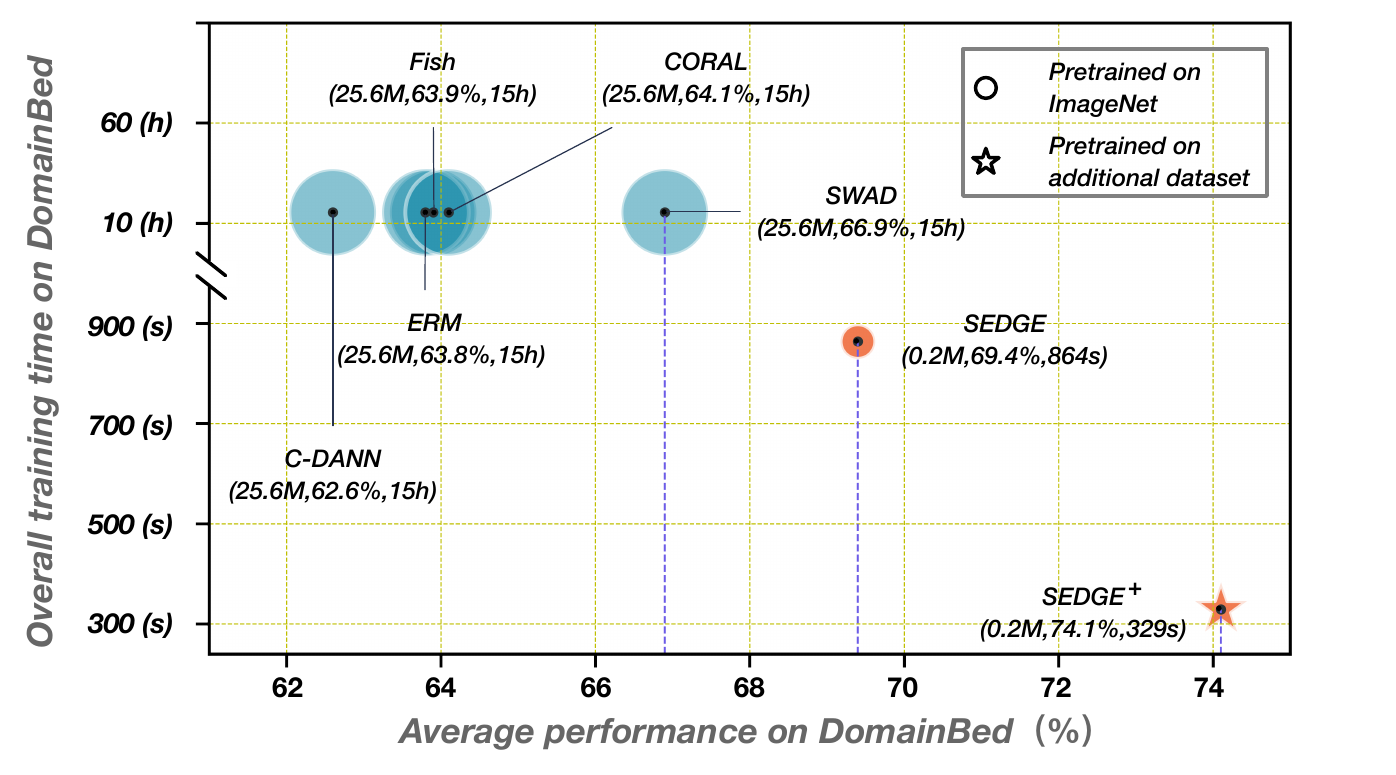}
  \end{center}

  \caption{The comparison of the average performance (x-axis, the higher the better) of different algorithms, their training time (y-axis, the smaller the better), and the number of their training parameters (the size of the marker). We also list the corresponding information (number of training parameters, test accuracy, training time) of each algorithm.
}

  \label{fig:cost-comparison}
\end{figure}

The advantages of our proposed learning paradigm lie in three aspects.
First, it shows a significant improvement
over the existing state-of-the-art (SOTA) result using the model pool pretrained only on ImageNet ~\citep{krizhevsky2012imagenet} dataset and 
gains even larger using the relatively larger model pool pretrained with additional datasets.
Second, it exhibits significantly higher training efficiency. The only parameters trained on the source domains contain (1) a linear adapter transforming model outputs to the target label space and (2) a lightweight ensemble network that has largely reduced the training cost on the source domains comparing with that fine-tuning the pretrained models.
We visualize the comparison of the performance w.r.t. to training parameter size and cost of training time in Figure~\ref{fig:cost-comparison}.
Last, this method illustrates a flexible way to utilize pretrained models, making it easier to exploit the abundant resource of pretrained models.

\section{Related work}\label{sec:related-work}

\subsection{Domain Generalization}
Mainstream domain generalization research can be divided into following categories.
(1) {Domain alignment}. In order to find the invariant representation across various domains, ~\citet{ganin2016domain} adversarially train a generator and discriminator to reach the equilibrium of optimal invariant features across domains, hence the classifier trained on multiple source domains would generalize well to target unseen domains. ~\citet{gong2019dlow} consider reducing domain discrepancy in a manifold space.
Some works resort to explicit feature distribution alignment on maximum mean discrepancy (MMD)~\citep{pan2010domain,tzeng2014deep,wang2018visual}, second order correlation~\citep{sun2016return,sun2016deep,peng2018synthetic}, moment matching~\citep{peng2019moment} and Wasserstein distance~\citep{zhou2020domain,lyu2021barycentric}, etc. Besides learning invariant representation, ~\citet{arjovsky2019invariant} consider to learn an optimal invariant classifier on top of the representation space, and enforce the learned classifier predicts according with causal mechanism.
(2) {Data manipulation}. ~\citet{tobin2017domain} first introduce this idea, which aims to create diverse training data to simulate unseen target domain. Besides, ~\citet{peng2018sim} and ~\citet{tremblay2018training} strengthen the generalization capability of the models via domain randomization, while other works consider using self-supervised learning~\citep{carlucci2019domain,kim2021selfreg} to match representation of an image with various augmentations. 
(3) {Meta-learning}. Inspired by~\citet{finn2017model} and with the expectation to capture the most transferable representations across domains, MLDG~\citep{li2018learning} split the multiple source domains data into meta-train and meta-test set to simulate domain shifts to learn more generalizable representations. 
~\citet{dou2019domain} introduce additional losses to explicitly pertain to the semantic structure in representations. 
~\citet{balaji2018metareg} consider learning a regularization function on classifier to avoid biasing to domain-specific information, while ~\citet{du2020learning} resort to regularize Kullback-Leibler (KL) divergence between distributions of latent representations within samples from different domains. 

While above categories more focus on algorithmic improvements, our proposed method SEDGE emphasizes the innovation of a learning paradigm based on a specialized pretrained model ensemble.

\subsection{Ensemble Learning}

Ensemble learning methods ~\citep{hansen1990neural,zhou2018diverse} exploit multiple models to produce prediction results and combine the results with various techniques, e.g., boosting ~\citep{schapire1990strength,freund1995boosting,moghimi2016boosted} or mean aggregation ~\citep{zhou2018diverse,zhang2020diversified}, etc., 
to achieve better performance than individual model alone.
These methods combine base model learning and ensemble as a whole and focus more on training diverse base models~\citep{zhou2018diverse}.

In DG, specifically, ensemble methods are used to exploit the relationship between source domains and the overall prediction results are composed of the superposition of the multiple networks on each domain. 
~\citet{mancini2018best} proposed to aggregate different predictions from specific trained source models. ~\citet{segu2020batch} proposed domain specific batch-norm statistics for each source domain. 
~\citet{zhou2021domain} proposed one shared CNN feature extractor with domain specific classifiers and each classifier is an expert to its own domain but non-expert to other domains. 
MulDEns~\citep{thopalli2021multi} relaxes the requirement for domain-specific models and uses a model-domain relevance matrix to define the relations between
models and 
domains.
Besides aggregating different domain expert models, other works consider combining model weights in different runs.
SWAD~\citep{cha2021swad} avoids overfitting models to local sharp minima by averaging model weights below a validation loss threshold. 
EoA~\citep{arpit2021ensemble} further lessens the frequent computations on validation set by averaging model weights simply from start to the end. 

These ensemble learning methods rely on training or fine-tuning from a pretrained model, share the same limitation of training cost and initialization model selection.
They did not consider the model specialty in different domains, classes or even samples.
We start from a novel perspective that incorporates various pretrained models without fine-tuning and builds a lightweight specialty-aware ensemble network, which illustrates better generalization performance and largely reduces training costs.

\section{Preliminaries}

\subsection{Problem Formulation}
Domain generalization aims to tackle the shift of data distribution among different domains by zero-shot transferring knowledge from seen to unseen domains.
Specifically, unlike domain adaptation, samples from unseen target domain(s) are inaccessible in domain generalization.
For a domain, its input and label space can be denoted as $\mathcal{X} \in \mathbb{R}^{d}$ and $\mathcal{Y} \in \mathbb{R}^{C}$, and its samples are observed constructing a dataset $D = \{ (\mathbf{x}_i, y_i) \}^{N}_{i=1}$ with $N$ sample points. 
Consider that we have $S$ source domains $\mathcal{D}_s = \left\{D_1, \dots, D_S\right\}$ and $T$ target domains $\mathcal{D}_t = \left\{D_1, \dots, D_T\right\}$ with different distributions on $\mathcal{X}\times \mathcal{Y}$ and sharing the label space.
Given instances drawn from source domains, the task is to learn a predictor parameterized by $\theta$ as $f_{\theta} \in \mathcal{M}$: $\mathbb{R}^{d} \longmapsto \mathbb{R}^{C}$, where $d$ is the dimension of input and $C$ is the number of classes in $\mathcal{Y}$.
We can define a population loss as $\mathcal{E}_{\mathcal{D}}(\theta)=\frac{1}{|\mathcal{D}|}\sum_{j=1}^{|\mathcal{D}|}\mathbb{E}_{\mathbf{x}_i \sim D_j} \left[ l(f_{\theta}(\mathbf{x}_i), y_i) \right]$ over the given domain $\mathcal{D}$.
The objective is to minimize the task-specific loss $l$ (e.g., cross-entropy loss 
for classification) over both source domains $\mathcal{D}_s$ and target domains $\mathcal{D}_t$ by only minimizing the empirical risk $\hat{\mathcal{E}}_{D_s}(\theta)$ w.r.t. model parameter $\theta$.
The performance on the target domains, then, measures the generalization ability of the learned model.

\begin{figure}[t]
    \begin{center}
    \includegraphics[width=1.0\textwidth]{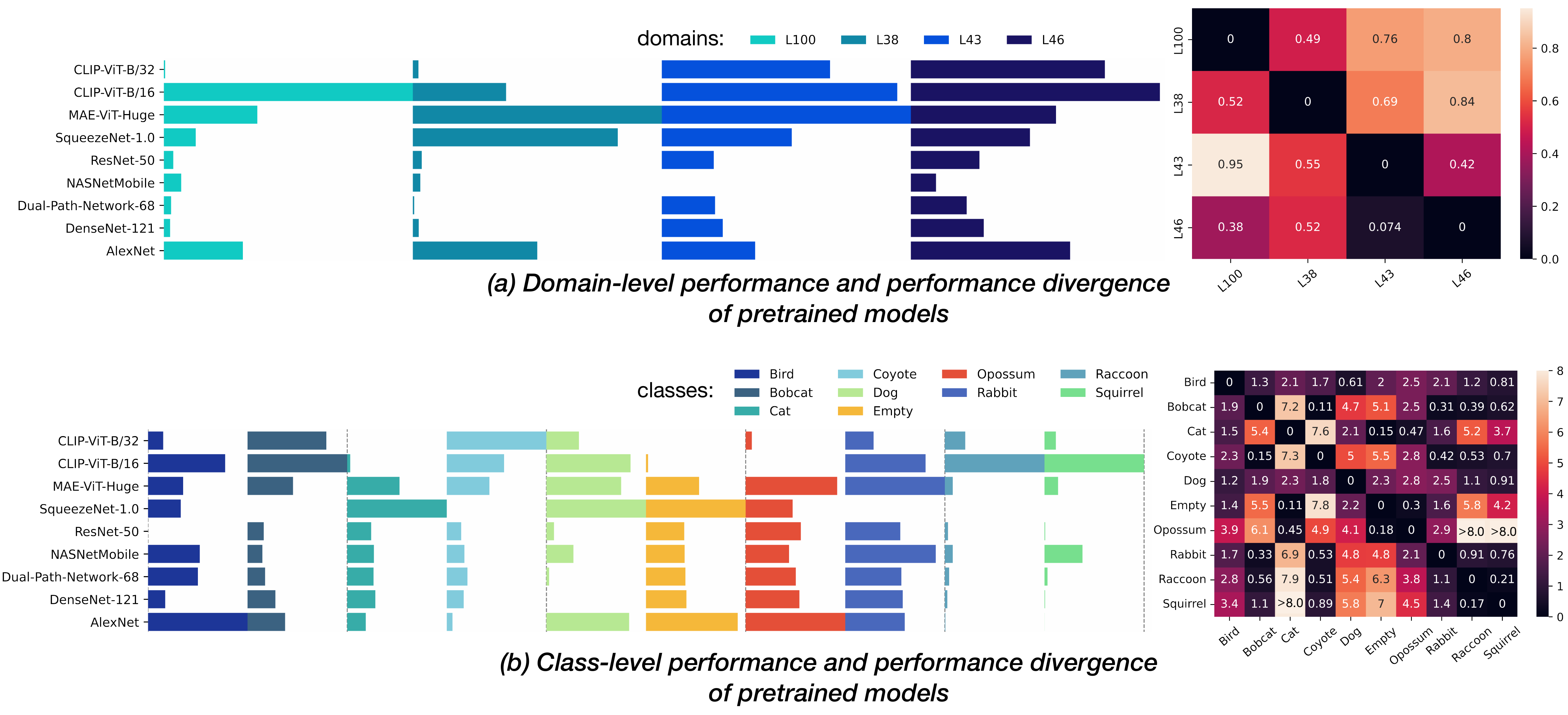}
  \end{center}

  \small
  \caption{
Performance distribution of the pretrained models over the samples within (a) different domains and (b) different classes.
Each column of the left panel displays the relative performance distribution of the pretrained models; the right panel shows the Kullback-Leibler divergence between the performance distribution of different (a) domains and (b) classes.
The comparison of the domain-level and class-level specialty
shows that the performance of the pretrained models differs more significantly at the finer level.
}
  \label{fig:spec different domains}
\end{figure}

\subsection{Preliminary Analysis}
\label{sec:preliminary_analysis}

In this section, we want to investigate the generalization ability of various pretrained models, to gain some insights to motivate our method.
As suggested by preliminary work~\citep{kumar2021fine}, a pretrained model with a linear probing layer (i.e., replacing the last layer of the pretrained model and retraining that) may achieve better accuracy in out-of-distribution scenarios than fine-tuning the whole model.
However, linear probing is not feasible
due to different pretrained models having different penultimate layer output feature dimensions. 
Instead, as shown in Figure~\ref{fig:training_paradigm} (B), we only train a label space adapter which learns the mapping function parameterized with $\phi$ as $h_{\phi} \in \mathcal{A}$: $\mathbb{R}^{C_o} \longmapsto \mathbb{R}^{C}$, where $C_o$ is 
the dimension of the label space
of the original pretraining dataset (pretraining domain).
Thus, all the pretrained models on the same pretraining dataset (e.g., ImageNet) share the same label adapter, through which it largely reduces the adaptation cost of the pretrained models on the new domains.

Given the pretrained model pool $\{f_{k}\}_{k=1}^K$ with $K$ pretrained models each of which is parameterized as $\theta_k$,
we further parameterize the adapted model $h_\phi(f_{k}(\cdot))$ as $\theta'_{k} = \left[\phi; \theta_k \right]$.
Then we train this shared adapter $h_\phi$ with the empirical loss $\hat{\mathcal{E}}_{D_s}(\phi)$ without fine-tuning the pretrained model parameters $\{\theta_k\}$.
With the trained adapter, we use the likelihood of the ground truth label $p(y_i\, \vert \, \mathbf{x}_i; \theta'_{k})$ on the $i$-th sample produced by each adapted model, which also indicates 
the confidence of the ground truth label $y_i$ of the model and $\sum_{y\in \mathcal{Y}}p(y\, \vert \, \mathbf{x}_i; \theta'_{k}) = 1$.
We utilize this likelihood as the evaluation metric of its sample-level model \textit{specialty}.

First, we analyze the specialty distribution of each pretrained model from an aggregation view (i. .e, domains and classes, respectively), and we verify if there exists a dominant pretrained model that generalizes best across different unseen domains.
We calculate domain-level model specialty as summation of logarithms of the sample-level specialty over all domain samples as $\sum_{(\mathbf{x}_i,y_i) \sim D} \log p(y_i\, \vert \, \mathbf{x}_i; \theta'_{k})$, on TerraIncognita~\citep{beery2018recognition} with four domains.
To reflect the relative model performance, we perform min-max normalization for model specialty values in the same domain.
These results are shown in Figure~\ref{fig:spec different domains} (a).
As can be seen, pretrained models vary greatly in performance on different domains, with no single model being dominant in all domains.
It suggests that finding a specific powerful pretrained model is non-trivial and not straight-forward for domain generalization.

Then, based on the previous finding, we further examine whether performance divergence also exists at a finer-grained level, such as class-level.
Similar as that at domain-level, Figure~\ref{fig:spec different domains} (b) presents the relative model performance on 10 classes in TerraIncognita-L100.
Model performance variances between classes are also noticeable.
To clearly compare model specialty differences at the two levels, we present heatmaps of specialty differences (measured by Kullback-Leibler divergence) for domain and class pairs, respectively in Figure~\ref{fig:spec different domains}.
The heatmaps exhibit a more pronounced divergence in model specialty at the finer class level.
It supports the necessity to utilize pretrained models on top of taking their fine-grained specialty, in finer-grained level even on each sample, into account.

\section{SEDGE: A New Paradigm for Domain Generalization}
In this section, we introduce our proposed learning paradigm, namely {\bf s}pecialized {\bf e}nsemble learning for {\bf d}omain {\bf ge}neralization (SEDGE), with the motivation and specific details of the whole method.
We first present the whole framework in Section~\ref{sec:framework} and then discuss the gathered pretrained models in Section~\ref{sec:pt-model-pool}.
After that, we introduce the model dispatcher with ensemble learning in Section~\ref{sec: model ensemble} and the corresponding learning algorithm in Section~\ref{sec:learning}.

\subsection{Framework: Pretrained Model without Fine-tuning}\label{sec:framework}

Recall that the focus of the paper is on leveraging pretrained models without fine-tuning to cope with domain generalization.
As motivated in Section~\ref{sec:preliminary_analysis}, each model has its own specialty and 
each sample may require to choose a specific set of models to give a good prediction. 
As a result, we learn the matching of pretrained models and samples from the source domains' training data.

\begin{figure}
    \begin{center}
    \includegraphics[width=0.9\textwidth]{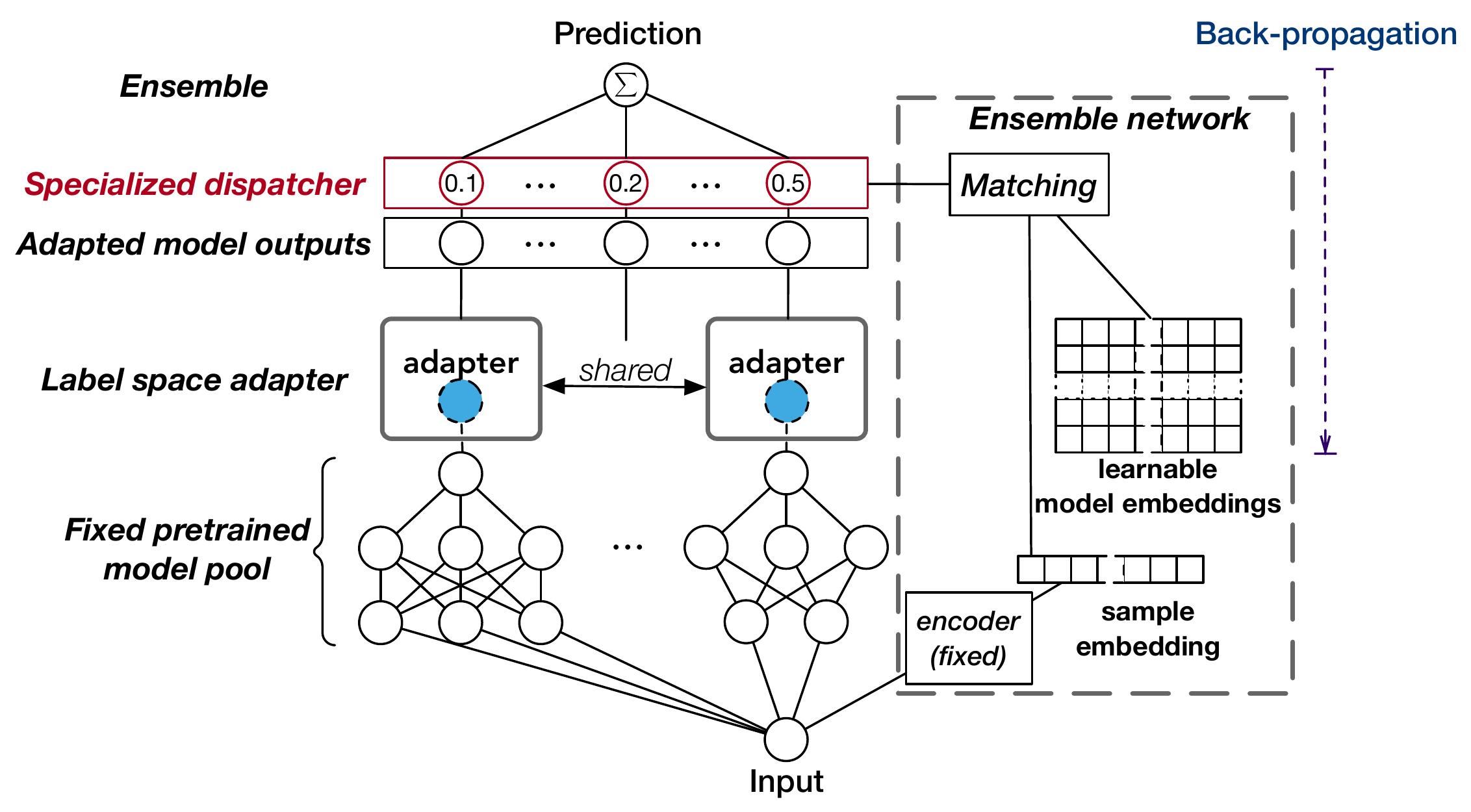}
  \end{center}
  \caption{\method framework. 
  Based on a pool of several fixed pretrained models, an ensemble network learns the matching of models and samples for model dispatching with the help of a label space adapter for prediction transformation.
  }

  \label{fig:paradigm}
\end{figure}

Based on this idea, as illustrated in Figure~\ref{fig:paradigm}, we propose a novel 
specialty-aware
domain generalization framework to dispatch an ensemble of specialized models for each sample.
Specifically, a label space adapter described in Section~\ref{sec:preliminary_analysis} is trained to transform the prediction of the pretrained models. 
And then, an ensemble network is learned to dispatch the models in a model pool to each sample according to their estimated specialty at sample level, and aggregate their outputs as an ensemble to output the final prediction for each sample.

\subsection{Pretrained Model Pool}\label{sec:pt-model-pool}

This section presents the pretrained models used in \method.
With more and more pretrained models being published,
it is straightforward to build a pretrained model pool consisting of several public pretrained models for direct adapting to novel domains.

On one hand, utilizing a ConvNet backbone~\citep{lecun2015deep} pretrained on ImageNet is a common practice of DG algorithms ~\citep{kim2021selfreg}.
Based on that, we first build Model Pool-A which only contains 15 models pretrained on ImageNet 
for fair comparison with the existing algorithms. 
In Model Pool-A, we incorporate the architectures including AlexNet (1)~\citep{krizhevsky2012imagenet}, DenseNet-121/169/201 (3)~\citep{iandola2014densenet}, Dual-Path-Network-68 (1)~\citep{chen2017dual}, NASNetMobile (1)~\citep{zoph2018learning}, ResNet-18/34/50 (3)~\citep{he2016deep}, SE-ResNet-50 (1)~\citep{hu2018squeeze}, SqueezeNet-1.0/1.1 (2)~\citep{iandola2016squeezenet}, and MAE-ViT-Base/Large/Huge (3)~\citep{he2021masked} with pretrained weights\footnote{\url{https://github.com/Cadene/pretrained-models.pytorch}}. 

On the other hand, several DG algorithms also use models pretrained on other datasets, such as 1G-1B~\citep{arpit2021ensemble} and ILSVRC12~\citep{thomas2021adaptive}. 
Therefore, we build Model Pool-B which contains two more CLIP models~\citep{radford2021learning}, ViT-B/16 and ViT-B/32, which are trained on a subset of the YFCC100M dataset of roughly the same size as ImageNet.
We denote \method using Model Pool-B as \methodplus$^{+}$ to distinguish it from the one using Model Pool-A.

Note that, the models pretrained on the same dataset, i.e., ImageNet, will share the same label adapter transforming the vanilla model outputs to the target label space.

\subsection{Ensemble Network}\label{sec: model ensemble}

As demonstrated by the findings in Section~\ref{sec:preliminary_analysis}, for generalizing to unseen domains, we need to take advantage of each of the pretrained models with consideration of their specialties.
Additionally, rather than using one model to predict each sample, we propose to use an ensemble of multiple pretrained models, which is known to bring less generalization error~\citep{ueda1996generalization}.
Moreover, our method incorporates sample-level model specialty into consideration and conducts fine-grained specialty-aware ensemble learning, which is novel comparing to the existing ensemble learning methods as discussed in Section~\ref{sec:related-work}.
This section will elaborate on the process of obtaining the most specialized pretrained models for a given sample  
and aggregating the outputs of selected models based on their specialty.  
The process can be divided into three steps as shown in Figure~\ref{fig:paradigm}.

First, we embed the input sample and the available models to a hidden space. 
For an image sample $\mathbf{x}_i$, we use a fixed pretrained model (i.e., ResNet-34 in our implementation) to embed $\mathbf{x}_i$ to $\mathbf{e}_i \in \mathbb{R}^{d_q}$. 
Meanwhile, we introduce a learnable latent variable $\mathbf{E}_m \in \mathbb{R}^{K\times d_m}$ as model embedding dictionary corresponding to the $K$ models $\{f_{k}\}_{k=1}^K$, which is randomly initialized and optimized during training.
Furthermore, we map $\mathbf{e}_i$ and $\mathbf{E}_m$ to a joint latent space as
\begin{equation}
    \mathbf{c}_i = \sigma(\mathbf{e}_i \mathbf{W}_i), \mathbf{C}_m = \sigma( \mathbf{E}_m \mathbf{W}_m),
\end{equation}
where $\mathbf{W}_i \in \mathbb{R}^{d_q\times d_v}$, $\mathbf{W}_m \in \mathbb{R}^{d_m\times d_v}$, and $\sigma(\cdot) = \max \{\cdot , 0\}$. 
We then perform matrix multiplication of $\mathbf{c}_i$ and $\mathbf{C}_m$ to calculate the matching score $\mathbf{s} = \mathbf{c}_i \mathbf{C}_m^T \in \mathbb{R}^{K}$ between the sample and each model.
To dispatch each model output to the final prediction on the sample,
we use one layer multi-layer perceptron and perform softmax operation to output the ensemble weights $\mathbf{w} = [w_1, \dots, w_K] \in \mathbb{R}^{K}$ with $w_k$ equals to
\begin{equation}
w_k = \dfrac{{e^{(\zeta (\mathbf{W}(\mathbf{s})))_k}}}{{\sum_{j=1}^K e^{(\zeta (\mathbf{W}(\mathbf{s})))_j}}},
\end{equation}
where $\mathbf{W} \in \mathbb{R}^{K\times K}$ and $\zeta(\cdot) = \log (1+ \exp (\cdot))$.
Finally, the prediction for $\mathbf{x}_i$ based on an ensemble of $K$ model outputs is written as
\begin{equation}
\hat{y}_i = \sum^{K}_{k=1} w_k   h_\phi(f_{k}(\mathbf{x}_i)),~ \text{ s.t. } \sum^{K}_{k=1} w_k = 1.
\end{equation}

\subsection{Learning Algorithm}\label{sec:learning}

As discussed above, the ensemble network acts as a model dispatcher through generating ensemble weights to aggregate multiple model predictions for each sample.
Section~\ref{sec:preliminary_analysis} shows that model performance varies significantly over samples.
Thus, we expect to assign more weights to the models with higher sample-level specialty to achieve the best utilization of the pretrained models.
That is, we try to minimize the estimation risk of the estimated model specialty on the ground truth,
i.e., $w_k$ and $p(y_i\, \vert \, \mathbf{x}_i; \theta' _{k})$, as
\begin{equation}
\begin{aligned}
\mathcal{L}_{c} = -\sum^{K}_{k=1} \left[ p\left( y_{i} \, \vert \, \mathbf{x}_{i}; \, \theta' _k \right) \cdot \ln (w_k) + (1-p\left( y_{i} \, \vert \, \mathbf{x}_{i}; \, \theta' _k \right)) \cdot \ln(1-w_k) \right].
\end{aligned}
\end{equation}
$\mathcal{L}_{c}$ is used to optimize the ensemble network to be a specialty-aware model dispatcher.

To train the general
label space adapter $h_\phi$ for all pretrained models, we incorporate the classification losses of adapted predictions of pretrained models
\begin{equation}\label{eq:l_b}
\begin{aligned}
\mathcal{L}_{b} = \sum^{K}_{k=1} w_{k}\cdot l\left( h_\phi(f_{k}(\mathbf{x}_i)), y_i\right),
\end{aligned}
\end{equation}
to update the shared adapter.
Additionally, we use the classification loss
\begin{equation}\label{eq:l_e}
\begin{aligned}
\mathcal{L}_{e} =  
l\left( \hat{y}_i, y_i \right)
\end{aligned}
\end{equation}
to optimize the likelihood of final ensemble output.
$\mathcal{L}_{e}$ is used to update both ensemble network and adapter.

It is worth noting that the only parameters to update
is the label space adapter and ensemble network, each of which is lightweight compared to the pretrained models which remain fixed in our method yet have been fine-tuned in the previous works.

\textbf{Relation to weight ensemble.}
Previous methods, such as SWAD~\citep{cha2021swad} and EoA~\citep{arpit2021ensemble}, show that averaging model weights during training can avoid overfitting and achieve better generalization performance. 
Their experimental results show superior performance compared with methods without weight averaging. 
While in \method, all pretrained model weights are not involved in training. 
Accordingly, we perform weight averaging for adapter and ensemble network starting from a certain iteration, which is served as a hyper-parameter.

\textbf{Top-$k$ model selection in inference.}
To save the inference time, we further select models with the highest $k$ ensemble weights and perform softmax on their ensemble weights for aggregation. 
In this paper, we set $k$ as 6. 

\section{Experiments}

\subsection{Evaluation Protocol} 

We conduct experiments on DomainBed suite ~\citep{gulrajani2020search}, which provides like-for-like comparisons between algorithms and has a standard evaluation protocol to follow.

\textbf{\textit{Datasets.}} 
We experiment on 5 real-world benchmark datasets including
PACS (4 domains, 9,991 samples, 7 classes)~\citep{li2017deeper}, VLCS (4 domains, 10,729 samples, 5 classes)~\citep{fang2013unbiased}, OfficeHome (4 domains, 15,588 samples, 65 classes)~\citep{venkateswara2017deep}, TerraIncognita (4 domains, 24,778 samples, 10 classes)~\citep{beery2018recognition}, and DomainNet (6 domains, 586,575 samples, 345 classes)~\citep{peng2019moment}. 

For fair comparison,
we follow the training and evaluation protocol of DomainBed~\citep{gulrajani2020search}. 
We use the \textit{training-domain validation set} protocol for model selection. 
Specifically, one domain in a dataset is selected as the target domain and the rest as source domains, from which 20\% of samples are used as the validation set. 
All runs are repeated 3 times using different random seeds, thus, with different train-validation splits.
The out-of-domain test performance averaged over all domains will be reported for each dataset.
In addition, we use the standard number of iterations of 5,000 for all datasets, with early-stop based on validated accuracy to reduce unnecessary computational costs.

\textbf{\textit{Baselines.}} 
We compare \method with some strong DG baselines including state-of-the-art.
As discussed in Section~\ref{sec:related-work}, some of the compared methods incorporate elaborate learning algorithms including
ERM~\citep{vapnick1998statistical}, CORAL~\citep{sun2016deep}, MLDG~\citep{li2018learning}, MMD~\citep{li2018domain}, DANN~\citep{ganin2016domain}, C-DANN~\citep{li2018deep}, and Fish~\citep{shi2021gradient}.

Some other works compared in our evaluation incorporate ensemble learning as listed as below.
\begin{itemize}
    \item [$\bullet$] 
    Stochastic Weight Averaging Densely (SWAD) ~\citep{cha2021swad}: SWAD performs weight ensemble during model training.
    \item [$\bullet$] 
    Ensemble of Average (EoA)~\citep{arpit2021ensemble}: 
    EoA combines both model ensemble and weight ensemble by taking an ensemble of moving average models from 6 runs. 
    They experiment with two different pretrained models as initialization.
    One is pretrained on ImageNet with ResNet-50 and the other is pretrained on both ImageNet and a much larger additional dataset, IG-1B, with a more advanced backbone, ResNeXt-50~\citep{xie2017aggregated}.
    We denote the latter one as EoA$^{+}$ to indicate it uses the additional dataset.
    \item [$\bullet$]
    Random ensemble: In contrast to \method of learning to select models for ensemble, we also compare it with average ensemble of $k$ models chosen randomly for each sample.
\end{itemize}
In addition, LP-FT~\citep{kumar2021fine} reveals the generalization of pretrained models and proposes an elaborated fine-tuning strategy. 
However, it does not follow the protocol of DomainBed and does not provide implementation details for replication. 
Our runs for LP-FT show it performs worse than ERM, whose results are shown in Appendix.
Note that, all the compared methods mentioned above incorporate a fine-tuning paradigm upon a pretrained model, which is essentially different to our method.

\subsection{DomainBed Benchmarking}

This section presents experimental results on the DomainBed suite, with performance comparison shown in Table~\ref{tab: experiment results} and training/inference time comparison in Table~\ref{tab:flops_param_comp}.

\textbf{Comparison with fine-tuning paradigm.}
The main difference between previous algorithms and \method lies in fine-tuning or no fine-tuning on top of pretrained models.
To verify whether the dispatcher of fixed pretrained models can outperform fine-tuning paradigm, we conduct a comparison of the algorithms that use models only pretrained on ImageNet, e.g., \method using Model Pool-A.
As shown in Table~\ref{tab: experiment results}, \method achieves an average performance of 69.4\%, exceeding SWAD by 2.5\%. 
Results show evidence that our novel paradigm is more effective than the traditional paradigm.

\textbf{Performance benefits from adding more pretrained models.}
\method provides a feasible way to incorporate the ever-emerging publicly available pretrained models.
Although 
Model Pool-A pretrained on ImageNet is in common use, ~\citet{kumar2021fine} finds that model pretrained on ImageNet may not be good for datasets such as DomainNet. 
By using Model Pool-B that includes models that have been pretrained on the CLIP dataset~\citep{radford2021learning}, \methodplus$^{+}$ further improves the average performance by 4.7\% over \method and ranks first on all datasets.
This confirms that \method paradigm is expected to generalize better on unseen domains by including models pretrained on more diverse datasets in the model pool.

\begin{table}
\centering
\caption{All baseline results are taken from their papers.
Our experiments are repeated 3 times using different random seeds.}

\resizebox{1.0\textwidth}{!}{
\begin{tabular}{l|l|l|l|l|l|l}
\hline
Algorithm           & PACS & VLCS          & OfficeHome    & TerraIncognita & DomainNet & avg. \\ \hline
\multicolumn{7}{c}{\textbf{\textit{Model Pool-A}}} \\ \hline
\specialrule{0em}{1pt}{1pt} \hline 
\multicolumn{7}{c}{Non-ensemble algorithms} \\ \hline
DANN (JMLR'16)~\citep{ganin2016domain} & 84.6$_{\pm 1.1}$ & 78.7$_{\pm 0.3}$ & 65.4$_{\pm 0.6}$ & 48.4$_{\pm 0.5}$ & 38.4$_{\pm 0.0}$ & 63.1 \\ \hline
CORAL (ECCV'16)~\citep{sun2016deep} & 86.0$_{\pm 0.2}$ & 77.7$_{\pm 0.5}$ & 68.6$_{\pm 0.4}$ &  46.4$_{\pm 0.8}$ & 41.8$_{\pm 0.2}$ & 64.1 \\ \hline
MLDG (AAAI'18)~\citep{li2018learning} & 84.8$_{\pm 0.6}$ & 77.1$_{\pm 0.4}$ & 68.2$_{\pm 0.1}$ & 46.1$_{\pm 0.8}$ &  41.8$_{\pm 0.4}$ & 63.6 \\ \hline
MMD (CVPR'18)~\citep{li2018domain} & 85.0$_{\pm 0.2}$ & 76.7$_{\pm 0.9}$ & 67.7$_{\pm 0.1}$ &  49.3$_{\pm 1.4}$ & 39.4$_{\pm 0.8}$ & 63.6 \\ \hline
C-DANN (ECCV'18)~\citep{li2018deep} & 82.8$_{\pm 1.5}$ & 78.2$_{\pm 0.4}$ & 65.6$_{\pm 0.5}$ & 47.6$_{\pm 0.8}$ & 38.9$_{\pm 0.1}$ & 62.6 \\ \hline
ERM (ICLR'21)~\citep{gulrajani2020search} &  85.7$_{\pm 0.5}$ & 77.4$_{\pm 0.3}$ & 67.5$_{\pm 0.5}$ & 47.2$_{\pm 0.4}$ & 41.2$_{\pm 0.2}$ & 63.8 \\ \hline
Fish (ICLR'22)~\citep{shi2021gradient} & 85.5$_{\pm 0.3}$ & 77.8$_{\pm 0.3}$ & 68.6$_{\pm 0.4}$ &  45.1$_{\pm 1.3}$ & 42.7$_{\pm 0.2}$ & 63.9 \\ \hline
\specialrule{0em}{1pt}{1pt} \hline 
\multicolumn{7}{c}{Ensemble algorithms} \\ \hline
SWAD (NIPS'21)~\citep{cha2021swad} & 88.1$_{\pm 0.1}$ & 79.1$_{\pm 0.1}$ & 70.6$_{\pm 0.2}$ & 50.0$_{\pm 0.3}$ & 46.5$_{\pm 0.1}$ & 66.9 \\ \hline
EoA (arxiv)~\citep{arpit2021ensemble} & \textbf{88.6} & 79.1          & 72.5          & 52.3           & \textbf{47.4}      & 68.0 \\ \hline 

random ensemble & 58.1$_{\pm 0.13}$ & 58.5$_{\pm 1.26}$  & 59.6$_{\pm 0.38}$ &  31.5$_{\pm 0.40}$  & 15.8$_{\pm 1.40}$  & 44.5 \\ \hline
\method         & 84.1$_{\pm 0.45}$  & \textbf{79.8}$_{\pm 0.12}$ & \textbf{79.9}$_{\pm 0.12}$ & \textbf{56.8}$_{\pm 0.21}$  & 46.3$_{\pm 0.39}$          &  \textbf{69.4}    \\ \hline
\specialrule{0em}{1pt}{1pt} \hline 

\multicolumn{7}{c}{\textbf{\textit{Model Pool-B}}} \\ \hline 
\specialrule{0em}{1pt}{1pt} \hline 

EoA$^{+}$ (arxiv)~\citep{arpit2021ensemble} & 93.2 & 80.4          & 80.2          & 55.2          & 54.6    & 72.7 \\ \hline 
random ensemble & 59.5$_{\pm 0.95}$  & 61.1$_{\pm 0.12}$   & 59.5$_{\pm 0.07}$ & 30.8$_{\pm 0.37}$  &  18.7$_{\pm 0.62}$    & 46.0 \\ \hline
\methodplus$^{+}$         & \textbf{96.1}$_{\pm 0.04}$  & \textbf{82.2}$_{\pm 0.03}$ & \textbf{80.7}$_{\pm 0.21}$ & \textbf{56.8}$_{\pm 0.29}$  & \textbf{54.7}$_{\pm 0.10}$          &   \textbf{74.1}   \\ \hline
\end{tabular}}
\label{tab: experiment results}
\end{table}

\textbf{Training cost comparison.}
\method only utilizes fixed pretrained models and learns to dispatch them through a lightweight ensemble network with the help of a linear label space adapter.
Therefore, the number of learnable parameters of \method (up to 0.6M) is minor compared with the normal image backbone network (25.6M for ResNet-50).
For fair training cost comparison, we run experiments of ERM, SWAD, \method on a single Nvidia Tesla V100 and compare their overall back-propagation time from the start of training to the end (or early-stop).
As shown in Table~\ref{tab:flops_param_comp}, training \method paradigm uses noticeably less time.
\methodplus$^{+}$ takes only 0.6\% of the time of ERM on DomainNet.
The significant training time advantage of the method and its surpassing performance suggest that \method is an effective and efficient paradigm for domain generalization.

\textbf{Inference cost comparison.}
As shown in Table~\ref{tab:flops_param_comp}, although ensemble methods like EoA and SEDGE achieve better generalization performance at the cost of higher inference FLOPs, SEDGE still manages to save a large amount of inference cost compared to the previous best ensemble model (half of the inference FLOPs compared to EoA).
This is because \method only selects models with the highest $k (<K)$ ensemble weights.
Therefore, only $k$ of $K$ models are activated for inference per sample, which reduces the inference cost to a large extent.

\begin{table}[]
\centering
\caption{The comparison of training and inference cost.
The run for SWAD on DomainNet failed due to out-of-memory. 
Here, ``\# parameters'' means the number of learnable parameters.
}
\resizebox{1.\textwidth}{!}{
\begin{tabular}{ll|l|l|l|l|l}
\hline
\multicolumn{2}{l|}{}                                           & ERM (our runs) & SWAD (our runs) & EoA (estimated) & \method (Pool-A) & \method (Pool-B) \\ \hline
\multicolumn{1}{l|}{\multirow{5}{*}{Training}} & PACS           & \multicolumn{1}{l|}{3.4h}                                & \multicolumn{1}{l|}{2.1h}                                 & \multicolumn{1}{l|}{20.4h}                                &    \multicolumn{1}{l|}{19.4s}                                                  &  \multicolumn{1}{l}{6.6s}                                               \\ \cline{2-7} 
\multicolumn{1}{l|}{}                          & VLCS           & \multicolumn{1}{l|}{3.6h} & \multicolumn{1}{l|}{2.4h} & \multicolumn{1}{l|}{21.6h}                                &  \multicolumn{1}{l|}{49.4s}   &  \multicolumn{1}{l}{8.3s}                              \\ \cline{2-7} 
\multicolumn{1}{l|}{}                          & OfficeHome     & \multicolumn{1}{l|}{3.3h}                               & \multicolumn{1}{l|}{2.1h}                                & \multicolumn{1}{l|}{19.8h}                               &  \multicolumn{1}{l|}{76.6s}                          &   \multicolumn{1}{l}{15.9s}                                                    \\ \cline{2-7} 
\multicolumn{1}{l|}{}                          & TerraIncognita & \multicolumn{1}{l|}{3.4h}  & \multicolumn{1}{l|}{2.1h}                                 & \multicolumn{1}{l|}{20.4h}                                &        \multicolumn{1}{l|}{46.4s}                                               &       \multicolumn{1}{l}{52.1s}                                                \\ \cline{2-7} 
\multicolumn{1}{l|}{}                          & DomainNet      & \multicolumn{1}{l|}{9.8h}                                 & \multicolumn{1}{l|}{/}                                  & \multicolumn{1}{l|}{58.8h}                                &       \multicolumn{1}{l|}{11.2m}    & \multicolumn{1}{l}{4.1m}                                         \\ \cline{2-7} 
\multicolumn{1}{l|}{}                          & \# parameters  & \multicolumn{1}{l|}{25.6M} & \multicolumn{1}{l|}{25.6M}  & \multicolumn{1}{l|}{153.4M}                               & \multicolumn{1}{l|}{0.2 $\sim$ 0.6M}                                     & \multicolumn{1}{l}{0.3 $\sim$ 0.6M}                                    \\ \hline
\multicolumn{1}{l|}{Inference}                 & GFLOPs     & \multicolumn{1}{l|}{3.9}                            & \multicolumn{1}{l|}{3.9}                              & \multicolumn{1}{l|}{23.5}                            & \multicolumn{1}{l|}{12.0}                            & \multicolumn{1}{l}{10.4} 
\\ \hline
\end{tabular}}
\label{tab:flops_param_comp}
\end{table}

\subsection{Ablation Study}\label{sec:ablation}

We want to verify the effectiveness of \method design by answering two research questions:
(\textbf{Q1}) Is grafting a label space adapter on top of model outputs sufficient, for utilizing pretrained models to generalize to novel domains?
(\textbf{Q2}) Is specialty-aware ensemble necessary, compared to an average ensemble method?

To verify whether a single model with an adapter can perform well, we train an individual adapter on source domains for each pretrained model in the model pool and compare their performance on target domain with state-of-the-art DG algorithm.
To show the ``cheating'' upper bound of performance under this ablation study, we report the \textit{best} single model performance on test set as \textit{best single model + adapter}.
As shown in Table~\ref{tab: ablation-adapter}, the \textit{best single model + adapter} among Model Pool-A can outperform SWAD only on OfficeHome.
It first indicates that the generalization ability of the fixed pretrained models may be more promising than model with fine-tuning on specific domains.
However, in other four datasets, it lags behind SWAD by a large margin.
This demonstrates a single pretrained model with a label space adapter is not sufficient to generalize to unseen domains, which motivates the main contribution of our method of introducing ensemble learning.

\begin{table} 
\centering
\caption{Results of applying label space adapter only and random ensemble. 
}
\resizebox{1.\textwidth}{!}{
\begin{tabular}{l|l|l|l|l|l|l}
\hline
Algorithm           & PACS & VLCS          & OfficeHome    & TerraIncognita & DomainNet & avg. \\ \hline
\multicolumn{7}{c}{\textbf{\textit{Model Pool-A}}} \\ \hline
SWAD (NIPS'21)~\citep{cha2021swad} & 88.1$_{\pm 0.1}$ & 79.1$_{\pm 0.1}$ & 70.6$_{\pm 0.2}$ & 50.0$_{\pm 0.3}$ & 46.5$_{\pm 0.1}$ & 66.9 \\ \hline
best single model + adapter & 79.7 & 73.6 & 78.3 & 49.2 & 32.5 & 62.7 \\ \hline
random ensemble & 58.1$_{\pm 0.13}$ & 58.5$_{\pm 1.26}$  & 59.6$_{\pm 0.38}$ &  31.5$_{\pm 0.40}$  & 15.8$_{\pm 1.40}$  & 44.5 \\ \hline
\method         & 84.1$_{\pm 0.45}$  & \textbf{79.8}$_{\pm 0.0}$ & \textbf{79.9}$_{\pm 0.12}$ & \textbf{56.8}$_{\pm 0.21}$  & 46.3$_{\pm 0.39}$          &  \textbf{69.4}    \\ \hline
\specialrule{0em}{1pt}{1pt} \hline 
\multicolumn{7}{c}{\textbf{\textit{Model Pool-B}}} \\ \hline   \hline 
best single model + adapter & 95.4 & 82.0 & 78.3 & 49.2 & 52.6 & 71.5 \\ \hline
random ensemble & 59.5$_{\pm 0.95}$  & 61.1$_{\pm 0.12}$   & 59.5$_{\pm 0.07}$ & 30.8$_{\pm 0.37}$  &  18.7$_{\pm 0.62}$    & 46.0 \\ \hline
\methodplus$^{+}$         & \textbf{96.1}$_{\pm 0.04}$  & \textbf{82.2}$_{\pm 0.03}$ & \textbf{80.7}$_{\pm 0.21}$ & \textbf{56.8}$_{\pm 0.29}$  & \textbf{54.7}$_{\pm 0.10}$          &   \textbf{74.1}   \\ \hline
\end{tabular}}
\label{tab: ablation-adapter}
\end{table}

DG algorithms that combine ensemble learning, such as SWAD and EoA, demonstrate promising performance.
A natural question is whether using an ensemble of pretrained models rather than a single model can improve performance. 
Following the ensemble approaches~\citep{lakshminarayanan2017simple} using mean average, we experiment a random ensemble over the fixed pretrained models, i.e., randomly sampling $k$ models for each sample and averaging their outputs for final prediction.
The results are shown in Table~\ref{tab: ablation-adapter}.
As can be seen, random ensemble results in worse performance than the single model, while \method with specialty-aware ensemble boosts the final performance significantly, albeit with strong or weak individual model performance, which verifies the necessity to select and ensemble the pretrained models based on their specialty over samples as mentioned in Section~\ref{sec:preliminary_analysis}  (\textbf{Q2}). 

\subsection{Further Analysis}

\begin{figure}
    \begin{center}
    \includegraphics[width=0.8\textwidth]{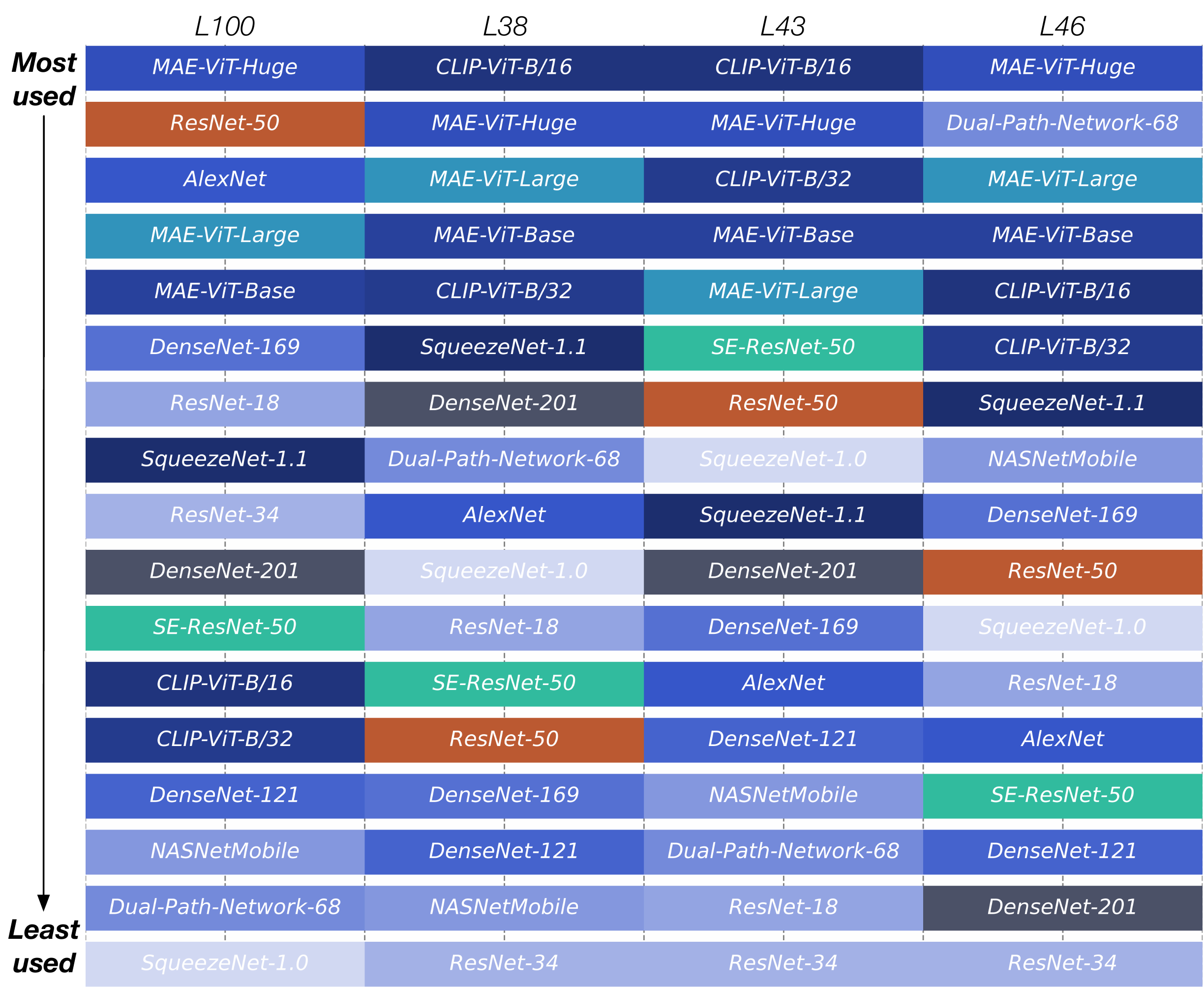}
  \end{center}
  \caption{Ranking models using the sum of ensemble weights on the sample in four domains of TerraIncognita. Each color block corresponds to a model. The higher rank indicates that this model is given a higher weight in predicting the samples in this domain.}
  \label{fig:model used count}
\end{figure}

As shown in Figure~\ref{fig:spec different domains}, model performance varies across domains, while \method is designed to dispatch specialized models for samples.
To analyze whether \method is handling as expected, we present its domain-level model assignment on different domains of TerraIncognita.
Specifically, we calculate the sum of ensemble weights assigned to a model as an evaluation of its importance.
In Figure~\ref{fig:model used count}, we show the rankings of model importance on different domains.
By comparing ranking between different domains, it can be seen that \method dispatches models quite differently over unseen target domains.
For example, while CLIP-ViT-B/32 model is used frequently on L38/43/46 datasets, it lags behind other models on L100.
It suggests that \method is making rational model selection as Figure~\ref{fig:spec different domains} shows CLIP-ViT-B/32 is not a powerful model on this domain.
Since the target domain is not known prior to making predictions, \method learns to find suitable models for each sample by learning on source domains only.

\section{Conclusions}

Domain generalization algorithms use the pretrained model as initialization for fine-tuning, while a few works have found that fine-tuning may lead to out-of-distribution performance degradation.
Different from the previous fine-tuning paradigm,
this paper proposes a novel paradigm for domain generalization, specialized ensemble learning which learns to dispatch an ensemble of fixed pretrained models for each sample based on the model specialty on it.
Experiments on five benchmark datasets show that our proposed method has achieved state-of-the-art performance with significant training cost reduction.

\bibliographystyle{plainnat}
\bibliography{main}

%%%%%%%%%%%%%%%%%%%%%%%%%%%%%%%%%%%%%%%%%%%%%%%%%%%%%%%%%%%%

\end{document}